\documentclass[journal]{IEEEtran}

\usepackage{amsmath,graphicx}
\usepackage{pdflscape}
\usepackage{tabularx}
\usepackage{lipsum}
\usepackage{xcolor}
\usepackage{amssymb}
\usepackage{multirow}
\usepackage{breqn}
\usepackage{afterpage,lipsum}
\usepackage{blindtext}
\usepackage[noend]{algpseudocode}
\usepackage[linesnumbered,lined,boxed, commentsnumbered]{algorithm2e}
\usepackage{blindtext}

\newcommand\blfootnote[1]{%
\begingroup
\renewcommand\thefootnote{}\footnote{#1}%
\addtocounter{footnote}{-1}%
\endgroup
}

\title{\LARGE \bf
Object Detection During Newborn Resuscitation Activities}

\author{\O yvind~Meinich-Bache,
        Kjersti~Engan,~\IEEEmembership{Senior Member,~IEEE,}
        Ivar~Austvoll,~\IEEEmembership{Member,~IEEE,}        
        Trygve~Eftest\o l,~\IEEEmembership{Senior Member,~IEEE,}   
        Helge~Myklebust,
        Ladislaus~Blacy~Yarrot, 
        Hussein~Kidanto and    
        Hege~Ersdal}  
       
\begin{document}

\maketitle
\thispagestyle{empty}
\pagestyle{empty}%

\begin{abstract}

\textit{Objective:} Birth asphyxia is a major newborn mortality problem in low-resource countries. International guideline provides treatment recommendations; however, the importance and effect of the different treatments are not fully explored. The available data is collected in Tanzania, during newborn resuscitation, for analysis of the resuscitation activities and the response of the newborn.  An important step in the analysis is to create activity timelines of the episodes, where activities include ventilation, suction, stimulation etc. 
\textit{Methods:} The available recordings are noisy real-world videos with large variations.  We propose a two-step process in order to detect activities possibly overlapping in time. The first step is to detect and track the relevant objects, like bag-mask resuscitator, heart rate sensors etc., and the second step is to use this information to recognize the resuscitation activities. The topic of this paper is the first step, and the object detection and tracking are based on convolutional neural networks followed by post processing. 
\textit{Results:} The performance of the object detection during activities were 96.97 \% (ventilations), 100 \% (attaching/removing heart rate sensor) and 75 \% (suction) on a test set of 20 videos. The system also estimate the number of health care providers present with a performance of 71.16 \%. 
\textit{Conclusion:} The proposed object detection and tracking system provides promising results in noisy newborn resuscitation videos. 
\textit{Significance:} This is the first step in a thorough analysis of newborn resuscitation episodes, which could provide important insight about the importance and effect of different newborn resuscitation activities.\\

\end{abstract}

\blfootnote{This work is part of the Safer Births project which has received funding from: Laerdal Global Health, Laerdal Medical, University of Stavanger, Helse Stavanger HF, Haydom Lutheran Hospital, Laerdal Foundation for Acute Medicine, University in Oslo, University in Bergen, University of Dublin – Trinity College, Weill Cornell Medicine and Muhimbili National Hospital. The work was partly supported by the Research Council of Norway through the Global Health and Vaccination Programme (GLOBVAC) project number 228203.}
\blfootnote{
\O, Meinich-Bache, K, Engan, I, Autsvoll and T, Eftest\o l is with the Dep. of Electrical Engineering and Computer Science, University of Stavanger, Norway.
H, Myklebust with Laerdal Medical, Norway.
L. B, Yarrot with the Research Institute, Haydom Lutheran Hospital, Manyara, Tanzania.
H, Kidanto with the School of Medicine, Aga Khan University, Dar es Salaam, Tanzania.
H, Ersdal with the Faculty of Health Sciences, University of Stavanger, Norway, and the
Department of Anesthesiology  and Intensive Care, Stavanger University Hospital, Norway.
}

\section{Introduction}
\label{sec:intro}

Globally, one million newborns die within the first 24 hours of life  each year. Most of these deaths are caused by complications during birth and birth asphyxia, and the mortality rates are highest in low-income countries \cite{birth}. As many as 10-20 \% of newborns require assistance to begin breathing and recognition of birth asphyxia and initiation 
of newborn resuscitation is crucial for survival \cite{birth, tactile, hege}.  International guidelines on newborn resuscitation exists, however,  the  importance and effect of the different treatments and therapeutic activities are not fully explored. 

Safer Births\footnote{www.saferbirths.com} is a research project to establish new knowledge on how to save lives at birth, and the project has, among other things, collected data during newborn resuscitation episodes at Haydom Lutheran Hospital in Tanzania since 2013.  
The collected data contains video recordings, ECG and accelerometer measurements from a heart rate sensor (HRS) attached to the newborn, and  measurements of pressure, flow and expired CO$_2$ from a bag-mask resuscitator (BMR).  
A thorough analysis of the collected data could provide important insight about different effects of the resuscitation
activities.
To be able to study such effects it is necessary to quantify the series of performed activities, in addition to measuring the condition of the newborn during resuscitation and knowing the outcome.   A timeline documenting activities like ventilation, stimulation and suction would be of immense value.  From such a timeline it would be possible to extract parameters like the amount of both total and continuous time used, the number of starts and stops for different activities etc.  The generation of the timelines should preferably be done automatically by using the collected signals and/or video, thus allowing large amounts of data to be analyzed.  The value of such timelines would clearly be i) for research and increased knowledge on the effects of newborn resuscitation activities. A future implementation of a complete system would also be useful on-site:  ii)  as a debriefing tool, summarizing the activities with no need to study video recordings and iii)  as a real-time feedback system. 

Previously, in Huyen et.al \cite{huyen}, our research group  
proposed an activity detector based on the HRS signals and the detector discriminated the activities  \textit{stimulation}, \textit{chest compressions} and \textit{other} with a accuracy of 78.7 \%.
Stimulation and chest compressions are therapeutic activities, whereas  \textit{other}  would include moving and drying the baby, touching the HRS etc.  These activities would result in movement in the HRS, and thus be visible in both the ECG and the accelerometer signals, but are not considered therapeutic activities or treatment of the newborn.  
Using automatic video analysis of the video recordings during the resuscitation episodes could potentially improve the performance  achieved using the HRS signals. Furthermore, video analysis could possibly detect activities and information that are difficult or impossible to detect from the ECG and accelerometer signals, like;  is the HRS attached to the newborn or not, and how many health care providers (HCPs) are present. 

The importance of video analysis of newborn resuscitation episodes has been well documented for both evaluation and  training purposes 
\cite{skaare,gelbart, silvia, nadler, kim}. However, manual inspection and annotation is very time consuming, and limits the amount of data that can be analyzed. In addition, a manual inspection entails privacy issues.
Thus, there is a need for automatic video analysis of these episodes.
Conventional image and pattern recognition methods, e.g segmentation and tracking,    has been applied in automatic video analysis for decades \cite{conventional}, but in recent years  Deep Neural Networks (DNNs) has shown it`s  superior strength in the field   
\cite{activityDetection1, activityDetection2,activityDetection3, activityDetection4}. 
In the topic of object and activity detection in resuscitation in general, others have propose the usage of passive radio-frequency identification (RFID) tags on the objects for object motion and interaction detection \cite{RFID1, RFID2, RFID3}.  Chakraborty et.al \cite{ActivityRecognitionResuscitation}  proposed an object and activity detector for trauma resuscitation video recordings based on object segmentation and a Markov Logic Network model. 
In the area of \textit{newborn} resuscitation 
Guo et.al \cite{activityDetectionNewbornVideos} proposed an activity detection system  for newborn resuscitation  videos based on DNN and linear  Support-Vector Machines (SVMs). Their dataset 
included 17 videos recorded with a frame rate of 25 frames per second (FPS) at a hospital in Nepal,
and the group aimed  to detect the activities \textit{stimulation}, \textit{suction},  \textit{ventilation} and \textit{crying}. 
The pre-trained \textit{Faster RCNN} network and the object class \textit{People} were used to propose areas involving the newborn, and motion salient areas were further used as input to two pre-trained Convolutional Neural Networks (CNN) from \cite{activityDetection1} designed to extract motion and spatial features. Further, the features was combined and used as input to linear SVMs, trained on their own dataset, to detect the activities.

All though there are similarities between the dataset from \cite{activityDetectionNewbornVideos} and our dataset, they are both noisy real-world videos with large variations, there are some specific tasks an challenges that differs between the studies. First, we aim to detect activities that are not newborn location dependent or movement dependent, like, the number of HCP present, and is the HRS attached or not. Second, in our dataset the newborns are wrapped in blankets most of the time, even before being placed at the resuscitation table, and the image examples from  \cite{activityDetectionNewbornVideos}, which shows fully uncovered newborns, are more infrequent in our dataset. Thus, using a pre-trained \textit{Person} detection network as suggested in  \cite{activityDetectionNewbornVideos} would most likely not be the best approach.  In addition, our videos are recorded with  variable frame rate, which in some case are very low and causes motion blurred images of poor quality, resulting in larger per frame motion variations than for images recorded with fixed frame rates. Considering all this, we believe that using an object detection and  tracking approach to localize the relevant activity detection areas would be a more robust first step  in activity detection.
Further, using the areas around each  objects would simplify the detection problem to a  binary classification problem for the specific activities; is the object being used in resuscitation or not. 
The topic of this paper is the first step and  the object detection and tracking is based on CNNs followed by post processing. Neural networks for object detection requires a lot of training data, 
so in addition to using  image frames from the videos, we use histogram matching \cite{hist} for  augmentation and also a synthetic dataset. The object detection is performed on each video frame and here we use the well known \textit{YOLOv3} \cite{YOLOv3} network, used in various object detection applications \cite{yolo1, yolo2, yolo3}. Post processing is used to fill in missing detections and track the area around the objects during the episodes.

\begin{figure*}[h]
\centering
\includegraphics[width=\textwidth]{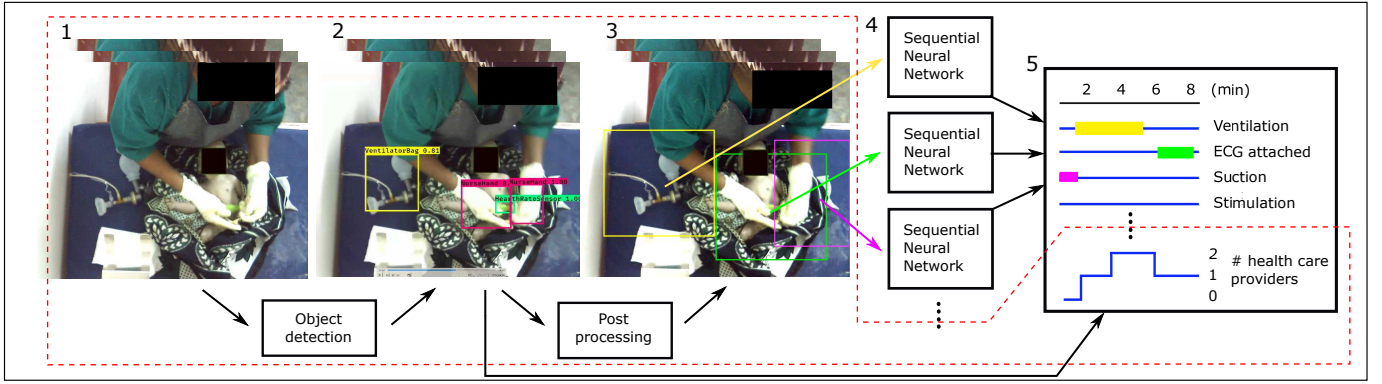}
\caption{Block scheme of the activity detection system.  The red dotted line encircles the steps proposed in this paper. 1: Generated dataset is input to YOLOv3 object detection network. 2: Detected objects. 3: Detected object area after post processing. 4: Sequence of images from areas are used as input to sequential neural networks. 5: Activity time lines is the final output. }  
\label{fig:oversikt}
\end{figure*}

\section{Data material}
\label{sec:data}

The dataset is collected using \textit{Laerdal Newborn Resuscitation Monitors} 
 (LNRM) \cite{huyenLNRM} and with cameras mounted over the resuscitation tables. The dataset contains almost 500 videos with corresponding LNRM data. The LNRM records the signals measured by the green
HRS and
the BMR, both shown at the top of Figure \ref{fig:SynthAndReal} C.

\begin{figure}[h]
\centering
\includegraphics[width=243pt]{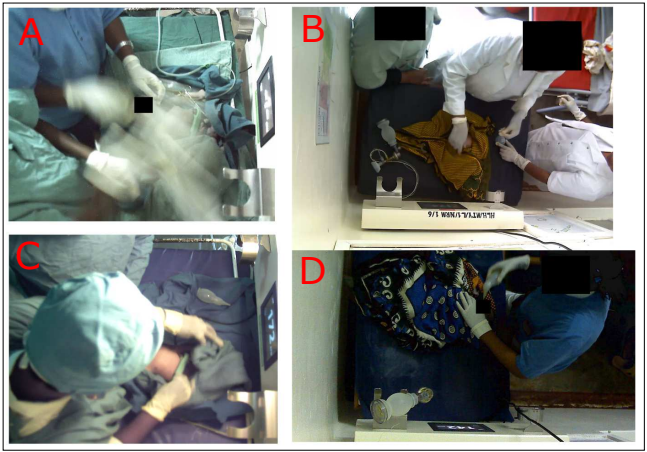}
\caption{A: Motion blurring due to low frame rate, 1024x1280. B: Camera far away, 1200 x 1600. C: Occlusion (ventilating newborn behind health care provider), 1024x1280. D: Poor lighting, 720 x 1280.  }  
\label{fig:var}
\end{figure}

The video recordings were initiated to provide additional support in cases and research objectives where the other collected signal or observed data were difficult to interpret. However, the videos are of variable
quality and camera and scene settings are not standardized for the different resuscitation tables included in the dataset. The variations are caused by different camera types, camera angles, video resolutions (1024$\times$1280, 720$\times$1280, and 1200$\times$1600), camera distances from resuscitation tables, variable frame rates (2-30 frames per second), unfocused cameras and light settings. All these variations, especially the variable frame rate, make automatic video analysis more challenging. In some cases the frame rate is as low as two frames per second, resulting in motion blurred image frames of poor quality.
In Figure \ref{fig:var} some of these challenges are depicted; 
A) Motion blurring, B) far away camera position, C) occlusion due to camera angle and D) poor lighting conditions.
In addition, the videos also have  variations like HCPs using different colored rubber gloves, HCPs that do not wear rubber gloves, different colored HCP uniforms and  clothing, and colorful and  patterned blankets brought by the mothers to wraps the newborn in.
The activity timelines that are relevant to generate are:
\begin{itemize}
\item 1) Bag-mask ventilations: Respiratory support.
\item 2) Suction: Removal of fluids from nasal and oral cavities using a device called suction penguin (SP).
\item 3) HRS attached to newborn or not.
\item 4) Stimulation: Warming, drying, and rubbing the newborns`s back.
\item 5) Chest compressions. Keep oxygenated blood flowing to the brain and other vital organs.
\item 6) Number of HCPs present.
\item 7) Newborn wrapped in blanket or not.
\end{itemize}
Activity 1), 2), 3), 4) and 5) can be detected by tracking the objects BMR, SP, HRS and HCPs hands (HCPH), and by analyzing their surrounding areas, 6) by counting the number of detected HCPH, and 7) by analyzing an area around the newborn, found from motion analysis and the location of the detected objects. 

\section{Methods}
\label{sec:methods}

A block scheme of the planned activity detection system is shown in Figure \ref{fig:oversikt}. The steps proposed in this paper is encircled with a red dotted line. These include dataset generation using the collected  videos, augmentation of images from the collected videos, generation of a synthetic dataset, object detection using YOLOv3 \cite{YOLOv3},  post processing to select the areas surrounding the relevant objects and an estimation of the number of HCPs involved in the resuscitation at each moment in time.

\subsection{Data Generation}
\label{dataGen}

A dataset, \textit{VideoD}, of 3093 images for object detection training is created by selecting evenly spread image frames from 21 randomly selected videos. The objects are manually labelled using the Image Labeler \cite{label}.

\subsubsection{Augmentation dataset}
\label{dataAug}

\begin{figure*}[t]
\centering
\includegraphics[width=\textwidth]{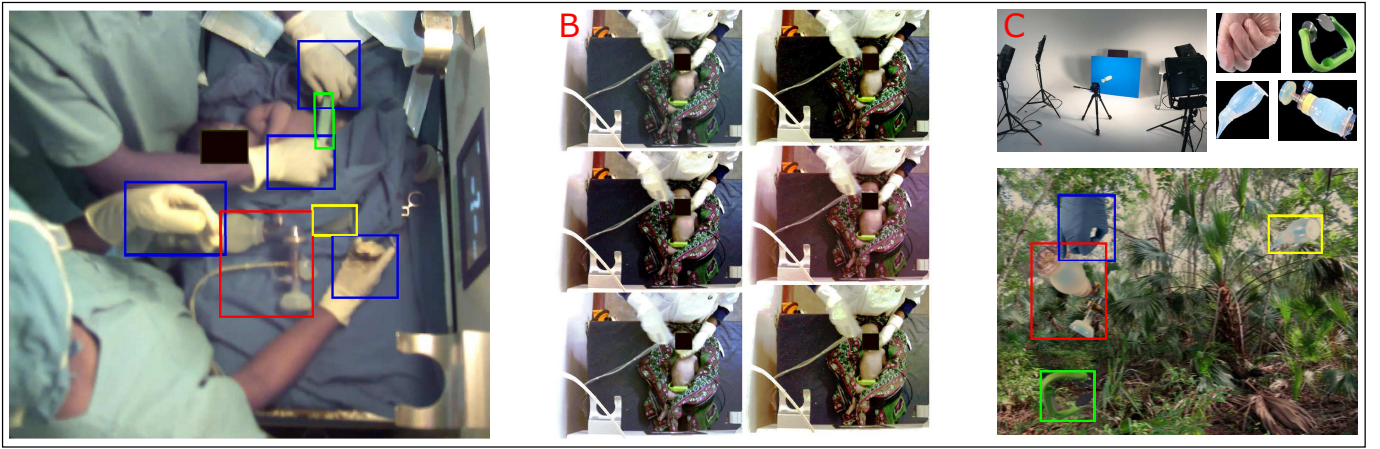}
\caption{A: Example of a frame used in \textit{VideoD}. B: Examples of histogram match augmented images from \textit{HistD}. C:  Scene for recording objects to be used in the generation of synthetic dataset, masked objects and an example of a generated frame in \textit{SynthD}. }  
\label{fig:SynthAndReal}
\end{figure*}

 \textit{VideoD} is further augmented to a new dataset, \textit{HistD}, by using histogram matching \cite{hist}. A frame from 10 randomly selected videos are used as histogram reference frames, and each of the images in \textit{VideoD} are augmented with each of the reference frames creating in total 34 023 images. 6 of 10 examples of the histogram match augmentation is shown for one of the frames in Figure  \ref{fig:SynthAndReal} B.

\subsubsection{Synthetic dataset}
\label{dataSynt}

A synthetic dataset, \textit{SynthD}, is  created in an attempt of generating example images with the variation found in the original dataset. Because of the colourful and patterned blankets used in the resuscitation videos, the objects we want to detect can appear on all kinds of backgrounds, thus over 6000 different backgrounds, both natural images and texture images are used. 
First, hands with different coloured gloves and no gloves, two types of BMR that both appear in the collected resuscitation videos, the HRS and the SP were video recorded in front of a blue screen in all possible angles. Object masks are created using  video frames, $I(x,y)_{i}$, where $x,y$ denote the pixel coordinates and $i$ the frame number, from the recorded object videos by:
\begin{equation}
\mathit{OM(x,y)_{i,c}} = I_B(x,y)_{i,c} - I_L(x,y)_{i,c} < T_{CK,c}
\end{equation}
where $c$ denote the object class, $I_B$ the blue channel, $I_L$ the RGB luminance value $(0.3 I_R + 0.59 I_G + 0.11 I_B)$ and $T_{CK}$ the chroma key thresholds for each $c$. Around 6300 masks per class are created in average. 

Next, a background is randomly drawn from the 6482 examples and objects and masks are cast at random positions onto the background. One example of each object, except from HCPH where we  use a number between one and three examples, is used. The objects are randomly scaled with the object`s typical size relative to the size of the image frame - found from \textit{VideoD}, and hue, saturation and lightness is also randomly chosen between 60-100 \% of the original object images. 

In order to make the object appear as realistic as possible, the final synthetic images are filtered with a small motion blur where the length, $len$, and angle, $\theta$, of the motion are randomly chosen. The scene for recording objects, masked objects and an example of a generated  synthetic image is shown in Figure, \ref{fig:SynthAndReal} C.

\subsubsection{Split image dataset}
\label{dataSplit}

In an attempt of better utilizing the resolution in the video frames and to be able to predict the smallest objects, the images in $HistD$ are split into five equally sized sub images generating a new dataset, $SplitD$. The four first images are generated from splitting the image into four parts, and the fifth is extracted at the center of the original image frame. This fifth sub image would typically contain more objects than the rest, and become an overlap of the other four sub images. The bounding box annotation is also split and the resulting bounding boxes is removed if they are $<$ 40 \% of the size of another box representing the same object in another sub image. This step ensures that all the resulting bounding boxes contain a significant part of the objects, making the resulting images good training examples.

\subsubsection{Dataset for testing}

A dataset, $TestD$, of 1000 images is created by selecting 50 evenly spread image frames from 20 randomly selected videos, not previously used for training,   where the mean duration per video is around 7 minutes. The test images are labelled using \textit{Image Labeler} \cite{label}. A split version, $TestD_{split}$, of $TestD$ is also created with the same procedure as explained in section \ref{dataSplit}.

\subsection{Object detection}

The proposed system uses the well known YOLOv3 \cite{YOLOv3} in the object detection step. YOLOv3 is comparable to the state of the art models on the mAP$_{50}$ metric \cite{YOLOv3}, and is chosen for the following reasons: 1) Speed - YOLOv3 can perform predictions on video streams in real time  - which could be useful in a future application for our proposed system, 2) YOLOv3 is state-of-the-art at predicting the correct class, rather than focusing on accurate bounding box predictions  - which suits the problem at hand well. 3) It predicts small objects with better precision than medium and large objects \cite{YOLOv3} - which also suits the problem at hand well,
and finally, 4)
due to the limited size of labelled training data, using transfer learning with a-state-of-the-art model as YOLOv3 as the starting point will most likely outperform any training from scratch. 

\subsubsection{Network structure (YOLOv3)}

YOLOv3 \cite{YOLOv3} is a fully convolutional network, meaning no fully-connected layers are used. It consist of 75 convolutional layers in total and performs downsampling by using convolutional layers with a stride of two instead of using pooling layers. The network also includes residual blocks \cite{residual} and  performs detection on three different scales in order to detect objects of different size. The detections on the different scales utilize feature maps from deeper layers in a similar concept  to feature pyramid networks \cite{pyramid} and the features go through convolutional layers before outputting 3D tensors with dimension:

\begin{equation}
N \times N \times [ 3 \times (4+1 + C)]
\end{equation}
where $N$ is the number of grids at that scale (13, 26 and 52 if image size is $416 \times 416$), 3 the number of bounding boxes for each grid, 4 the box coordinates and size, 1 the objectness prediction, $oP$, and C the number of object classes. The YOLO algorithm further performs non-maximum suppression: Removing predicted object with an objectness score below a threshold, $T_o$, and by removing predictions of same class where the bounding box overlap more than threshold $T_{IoU}$.

\subsubsection{Post processing object detection}
\label{PP}

\begin{figure}[h]
\centering
\includegraphics[width=240pt]{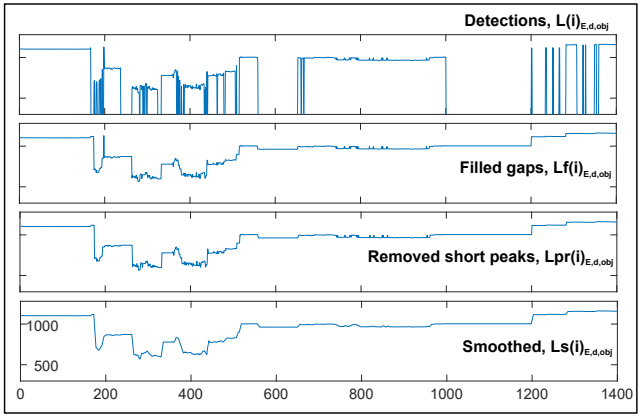}
\caption{Example of post processing  the centroid X-coordinate of the detected bag-mask resuscitator (BMR). Horizontal axis is the image frame in the video and vertical axis the pixel position in the frame. }  
\label{fig:eventTracking}
\end{figure}

Post processing is performed on the detection of \textit{BMR}, \textit{SP} and  \textit{HRS}
to fill in missing detections in frames and to create areas surrounding the object throughout the video. 
Since we can have multiple true occurrences of  HCPH in the same frame, HCPH do not undergo these steps. Denote
$obj \in {1:4}$ to be the object classes  where $1=$ \textit{BMR}, $2=$ \textit{SP}, $3=$ \textit{HRS} and $4=$ \textit{HCPH}, and $N_{E,i}$ to represent the number of detections in image, i, of episode, E.
For  $obj_{p} \in \{1,2,3\}	\subset obj$ 
we estimate the most likely object position in each $i$ by; first, creating blank images, $IB(x,y,obj_p)_{E,i}$. 
Second, for each pixel areas, $pA_{E,i,obj_{p,n}} = \{x^{E,i,obj_p}_{n}, y^{E,i,obj_P}_{n}\}$, representing all pixel coordinates of a   detected object,  $obj(n)_{E,i}$, in an image we add the detection`s $oP$ score, $oP(n)_{E,i,obj_{P}}$, to the matching coordinates in $IB(x,y,obj_{p})_{E,i}$. \\
\\
\emph{For} $n=1:N_{E,i}$ \emph{do}:
\begin{equation}
\mathit{IB(x,y,obj_{p})_{E,i}} = \left\{ 
  \begin{array}{l l}
    \mathit{IB(\cdot)} + oP(n)_{E,i,obj_{p}},\\ 
     \quad \quad \forall \{x,y\} \in  pA_{E,i,obj_{p,n}(n)} \\
      \quad \quad \text{if} \quad obj(n)_{E,i}=obj_{p} \\
    \mathit{IB(\cdot)},  \quad  \text{otherwise}
  \end{array}
\right .
\label{eq:diffIm}
\end{equation}
Further the centroid coordinates, $(x_c^{{E,i,obj_p}},y_c^{E,i,obj_p})$, of the most likely object position is found from:
\begin{equation}
(x_c^{(\cdot)},y_c^{(\cdot)}) = cent(max(IB(x,y,obj_p)_{E,i}> T_{obj_p})
\end{equation}
where  $T_{obj_p}$ defines thresholds for the different object classes.  
Denote $d \in {X,Y}$. Each  $x_c^{(\cdot)}$ and $y_c^{(\cdot)}$ are stored in location vectors, $L(i)_{E,d,obj_p}$, representing timelines of the center position of each object as a function of the video frames. $L(i)_{E,d,obj_p}$ further undergoes the three post processing steps illustrated with an example in Figure \ref{fig:eventTracking}, listed as follows:\\
1) Filling detection gaps by choosing the previous detected value $\rightarrow \quad Lf(i)_{E,d,obj_p}$. \\
2) Short peak removal. If  $||Lf(i)_{(\cdot)} -  Lf(i-1)_{(\cdot)} || > T_{peak}$, we check if it is an actual large change in object position, or if it returns to a value where\\ $Lf(i+1:i+10)_{(\cdot)} -  Lf(i-1)_{(\cdot)} || <T_{stable}$. 
This step filters out short false detections of the objects, and outputs the peak removed signal, $Lpr(i)_{E,d,obj_p}$. \\
3) Signal smoothing by applying a moving average filter of length $N_{f1}$: 
\begin{equation}
\mathit{Ls(i)_{E,d,obj_p}} = \frac{1}{N_{f1}}\sum_{l=-N_{f1}/2}^{N_{f1}/2}{\mathit{Lpr(l)_{E,d,obj_p}}}
\end{equation}

Finally, object area tracking throughout sequences is  performed by adding a $500 \times 500$ bounding box, $BB_{track, E,obj_p}$, around each $\mathit{Ls(i)_{E,d,obj_p}}$ onto the original videos. The size of  $BB_{track, E,obj_p}$  ensure that it is possible to detect what activities are performed in the area, and thus discriminate the activities from movement and noise. An example of the tracking results is shown in step 3 of Figure \ref{fig:oversikt}.

\subsection{Estimation of number of health care providers present}
\label{nurse}
Timelines of the number of HCPs present in the resuscitation videos are generated from the number of detected hands in the image frames, $\mathit{nH(i)_{E}}$.\\
\\
\emph{For} $n=1:N_{E,i}$ \emph{do}: 
\begin{equation}
\mathit{nH(i)_{E}} = \left\{ 
  \begin{array}{l l}
    nH(i)_{E} + 1, \quad  \text{if } obj(n)_{E,i} = 4 \\
   \quad  \quad  \quad  \text{and } oP(n)_{E,i} > T_{\mathit{HCPH}} \\
    nH(i)_{E},  \hspace{0.8cm} \text{otherwise}
  \end{array}
\right .
\label{eq:diffIm}
\end{equation}
where $T_{\mathit{HCPH}}$ is a threshold for detection of HCPHs.
To remove noise, $nH(i)_{E}$ is further smoothed by a moving average filter:

\begin{equation}
\mathit{\overline{nH}(i)_{E}} = \frac{1}{N_{f2}}\sum_{l=-N_{f2}/2}^{N_{f2}/2}{\mathit{nH(l)_{E}}}
\end{equation}
where $N_{f2}$ is the filter size. Finally, $\mathit{\overline{nH}(i)_{E}}$ is converted to the detected  number of HCPs, $nHCP(i)_{E}$,  by:

\begin{equation}  \label{adi}
 \mathit{nHCP(i)_{E}} = 
  \begin{cases} 
  0 & \text{if } \quad \mathit{\overline{nH}(i)_{E}} \leq T_{zero} \\
  1 & \text{if } \quad  T_{zero} < \mathit{\overline{nH}(i)_{E}}  \leq T_{one}\\
  2 & \text{if } \quad  T_{one} < \mathit{\overline{nH}(i)_{E}} \leq T_{two}\\
  3 & \text{if } \quad  \mathit{\overline{nH}(i)_{E}} > T_{two}
  \end{cases}
\end{equation}

\section{Experiments}

We used the original pretrained weights for YOLOv3, \textit{darknet53}, and trained different models by further training the weights with four different sets of training data, $VideoD$, $HistD$, $HistD+SynthD$ and $SplitD + SynthD$. An initialization stage is used to get a stable loss by first freezing all layers except the top 3 layers. In the next and final stage all layers are further trained with learning rate decay and early stopping. The batch size was set to 16.
The mean Average Precision (mAP) criterion defined in the PASCAL VOC 2012 competition\footnote{http://host.robots.ox.ac.uk/pascal/VOC/voc2012/} was used to compare  single-image object detection results from  the models trained on the four different mixtures of the datasets. 
mAP is a function of \textit{precision}, \textit{recall} and the Intersection over Unions (IoU), the overlap between predicted and true bounding box.  The threshold for IoU was set to 0.5. 

The best models were further used in detection of the objects and the post processing steps to evaluate the performance of the proposed object regions. The proposed regions were added to the original video and the detection results were manually evaluated by annotating timelines using the video annotation tool ELAN\footnote{https://tla.mpi.nl/tools/tla-tools/elan/}. The  annotated timelines for each $E$ are:
\begin{itemize}
\item The number of HCPs: $\mathit{nHCP_{ref,E}(i)}$, 
\item activities - ventilations, attaching or removing HRS,  and suction, $A_{obj_p,E}(i)$,
\item is the object visible: $V_{obj_p,E}(i)$ and
\item is the object detected: $D_{obj_p,E}(i)$ ($>$ half the object is included in $BB_{track, E,obj_p}$)
\end{itemize}

The main task of the object detection and tracking is to find approximate regions around the objects that can be used  for further activity recognition. The aim is not to propose very accurate regions that centers the object perfectly, but more importantly to propose smoothly updated regions  that surround the object over time. Thus, we classify a tracking result as correct if the object is at least 50 \% included in the proposed region.

Since our aim is to track a single object of each of the classes \textit{SP}, \textit{HRS} and \textit{BMR} throughout the whole video, we can evaluate the objects individually. The established metric Multiple Object Tracking Accuracy (MOTA) can be seen in the context of single-object short-term tracking
and be simplified to the percentage of correctly tracked frames \cite{MOTAsimple}. Thus, the performance, P, is evaluated for each object class and each episode, E, by the general equation

\begin{equation} \label{P}
 P = (\frac{1}{N_s}  \sum_{i=1}^{N_s}{I_f(i)}) *100
\end{equation}
where $N_s$ is the number of frames in the  episode and $I_f(i)$ an indicator function defined as 1 if $|detection(i)_E - reference(i)_E| = 0$ and 0 otherwise.
The average performance, $\overline{P}$, of the post processed object detection are estimated using Eq. \ref{P} 
with $D_{obj_p,E}(i)$  as \emph{detection} $V_{obj_p,E}(i)$) as \emph{reference}, and by averaging over the episodes. 

Further, we evaluate the performance of the object detection during the relevant resuscitation activities, ventilation (BMR), Attaching or removing HRS and suction (SP). From
$A_{obj_p,E}(i)$ we locate the activity sequences and use them as \emph{reference} in Eq. \ref{P}. Their corresponding sequences in time in $D_{obj_p,E}(i)$ is here used as \emph{detection} and
an activity is classified as detected if the detection overlap with the reference data $>$ 80 \% of the time.

The timelines $\mathit{nHCP(i)_{E}}$ is found as explained in Section \ref{nurse} and the average performance, $\overline{P}$, of the prediction of number of HCPs is estimated 
using Eq. \ref{P} with $\mathit{nHCP_{ref,E}(i)}$ as \emph{reference} and  $\mathit{nHCP(i)_{E}}$ as \emph{detection}. In addition, the average prediction error,  $\overline{E}$, of $||\mathit{nHCP_{ref,E}(i)} - \mathit{nHCP(i)_{E}}||$ is estimated over the episodes.
The total  performance, $P$,  of the classes \textit{no HCP}, \textit{one HCP}, \textit{two HCP} and \textit{three (or more) HCP} is also estimated using Eq. \ref{P}, where the class-relevant sequences in $\mathit{nHCP_{ref,E}(i)}$ is the \emph{reference} and the corresponding sequences in time in $\mathit{nHCP(i)_{E}}$ is the \emph{detection}.

When the results are averaged over results from individual episodes, quartile measurements, $Q$, are also provided.

The experiments are done using Python\footnote{https://www.python.org/} and a Keras\footnote{https://keras.io/} implementation of YOLOv3 developed by user \textit{qqwwee}\footnote{https://github.com/qqwweee/keras-yolo3} with minor modifications. 
Since the objects often are occluded in the videos and the camera distance varies, the objects`s size and form have large variations. Therefore, we have chosen to use the YOLOv3 anchor boxes determined using k-means clustering on the large COCO dataset \cite{YOLOv3} instead of estimating anchor boxes from our limited truth data.

The threshold and parameter values used in the experiments are: $T_{CK,c} \in \{80, 180\}$, $len=3-7$, $\theta=3-10$, $T_o=0.05$, $T_{IoU}=0.45$ , $T_{obj} = [0.1, 0.05, 0.1]$ for BMR, SP and HRS, $T_{\mathit{HCPH}} = 0.1$ $T_{peak} =200$, $T_{stable} =50$, $T_{zero}=0.2$, $T_{one}=2$, $T_{two}=4$, $N_{f1}= 5$ and $ N_{f2} = 40 $.
\label{sec:ex}

\section{Results}
\label{sec:results}

The mean average precision, mAP, results are listed in Table \ref{tab:table1} 
for the object detection using models trained on the datasets $VideoD$, $HistD$, $HistD+SynthD$ and $SplitD + SynthD$. For the objects HCPH, BMR and HRS using a combination of $HistD$ and $SynthD$ and image size $416 \times 416$ provided the best results. There was no significant improvement  by increasing the image input size to $608 \times 608$.
For detection of SP we achieved the best result by using a model trained on $SplitD$ and $SynthD$, and an image size of $608 \times 608$. This model also provided the best overall  mAP.

\begin{table}[h]
\centering
\caption{Object detection results, measured with mAP$_{50}$, for models trained with different  datasets. 
HCPH = health care provider hand, BMR = bag-mask resuscitator, HRS = heart rate sensor and SP = suction penguin.} 

\begin{tabular}{l|c|c|c|c|}
\cline{1-5}
\multicolumn{1}{|c|}{}    & \multicolumn{1}{l|}{\textbf{\begin{tabular}[c]{@{}l@{}}$VideoD$ \\$ 416\times416$\end{tabular}}
} & \multicolumn{1}{l|}{\textbf{\begin{tabular}[c]{@{}l@{}}$HistD$ \\$ 416 \times 416$\end{tabular}}
} & \multicolumn{1}{l|}{\textbf{\begin{tabular}[c]{@{}l@{}}$HistD + $ \\$ SynthD$ \\$ 416 \times 416$\end{tabular}}
} & \multicolumn{1}{l|}{\textbf{\begin{tabular}[c]{@{}l@{}}$SplitD + $ \\$ SynthD$ \\$ 608 \times 608$\end{tabular}}
} \\ \hline
\multicolumn{1}{|l|}{\textbf{HCPH}}       			& 63.91			& 68.49			& {\textbf{70.07}}	& 68.55                               \\ \hline
\multicolumn{1}{|l|}{\textbf{BMR}}   				& 57.45         & 57.54         & {\textbf{62.07}} & 59.77                                                    \\ \hline
\multicolumn{1}{|l|}{\textbf{HRS}} 					& 62.79         & 71.61         & {\textbf{79.38}} & 73.49                                                    \\ \hline
\multicolumn{1}{|l|}{\textbf{SP}}  					& 25.92         & 18.86         & 19.25                                 & {\textbf{42.02}}                    \\ \hline
\multicolumn{5}{|l|}{}                                                                                                                                                                                                                            \\ \hline
\multicolumn{1}{|l|}{\textbf{Total}}       			& 52.52         & 54.12         & 57.69                                 & {\textbf{60.96}}                    \\ \hline

\end{tabular}
\label{tab:table1}
\end{table}

\begin{table}[h]
\caption{Performance results. Top section: Object detection (using a 500x500 area) after post processing. Middle: object tracking when relevant activities occurs (\# detected / \# true). Bottom: Prediction of the number of health care providers.}  
\centering
\begin{tabular}{lcc}

\cline{1-3}
\hline
\multicolumn{1}{|c|}{\textit{\textbf{\begin{tabular}[c]{@{}l@{}}Object detection \hspace*{0.09cm} \\ (post processed)\end{tabular}}}} & \multicolumn{1}{c|}{\textbf{\begin{tabular}[c]{@{}c@{}}$\overline{P}$ \end{tabular}}} 	& \multicolumn{1}{c|}{\textbf{Q  (25,50,75)}} \\ \hline
\multicolumn{1}{|l|}{\textbf{BMR}} 		& \multicolumn{1}{c|}{96.66 \%} 	& \multicolumn{1}{c|}{\begin{tabular}[c]{@{}c@{}}96.23, 100, 100 (\%)\end{tabular}} \\ \hline
\multicolumn{1}{|l|}{\textbf{HRS}} 		& \multicolumn{1}{c|}{97.88 \%} 	& \multicolumn{1}{c|}{100, 100, 100 (\%)} \\ \hline
\multicolumn{1}{|l|}{\textbf{SP}} 		& \multicolumn{1}{c|}{76.86 \%} 	& \multicolumn{1}{c|}{\begin{tabular}[c]{@{}c@{}}70.99, 81.67,  92.82 (\%)\end{tabular}} \\ \hline
{\textbf{}} &  &  \\ \hline
\multicolumn{1}{|l|}{\textit{\textbf{\begin{tabular}[c]{@{}l@{}}Object detection \\ during activity\end{tabular}}}} & \multicolumn{1}{c|}{\textbf{$P$}} & \multicolumn{1}{c|}{\textbf{Activities}} \\ \hline
\multicolumn{1}{|l|}{\textbf{BMR}} 	& \multicolumn{1}{c|}{96.97 \% (64/66)} 			& \multicolumn{1}{c|}{Ventilation} \\ \hline
\multicolumn{1}{|l|}{\textbf{\begin{tabular}[c]{@{}l@{}}HRS\end{tabular}}} 	& \multicolumn{1}{c|}{100 \% (43/43)} & \multicolumn{1}{c|} {Attach/remove HRS} \\ \hline
\multicolumn{1}{|l|}{\textbf{SP}} 		& \multicolumn{1}{c|}{75.00 \% (45/60)} 		& \multicolumn{1}{c|}{Suction} \\ \hline
{\textbf{}} &  &  \\ \hline

\multicolumn{1}{|l|}{\textit{\textbf{\begin{tabular}[c]{@{}l@{}}HCP detection \end{tabular}}}} & \multicolumn{1}{c|}{\textbf{$P$}} & \multicolumn{1}{c|}{\textbf{}} \\ \hline
\multicolumn{1}{|l|}{\textbf{No HCP}} 		& \multicolumn{1}{c|}{90.70 \%} 				& \multicolumn{1}{c|}{} \\ \hline
\multicolumn{1}{|l|}{\textbf{One HCP}} 		& \multicolumn{1}{c|}{90.48 \%} 				& \multicolumn{1}{c|}{} \\ \hline
\multicolumn{1}{|l|}{\textbf{Two HCPs}} 	& \multicolumn{1}{c|}{53.31 \%} 				& \multicolumn{1}{c|}{} \\ \hline
\multicolumn{1}{|l|}{\textbf{\begin{tabular}[c]{@{}l@{}}Three (or more)\\ HCPs\end{tabular}}} & \multicolumn{1}{c|}{6.88 \%} & \multicolumn{1}{c|}{} \\ \hline

\multicolumn{1}{|l|}{\textit{\textbf{\begin{tabular}[c]{@{}l@{}} \end{tabular}}}} & \multicolumn{1}{c|}{\textbf{{$\overline{P}$}}} & \multicolumn{1}{c|}{\textbf{Q  (25,50,75)}} \\ \hline
\multicolumn{1}{|l|}{\textbf{\begin{tabular}[c]{@{}l@{}}HCP correct pred.\end{tabular}}} & \multicolumn{1}{c|}{71.16 \%} & \multicolumn{1}{c|}{\begin{tabular}[c]{@{}c@{}}50.72, 78.56, 89.45 (\%)\end{tabular}} \\ \hline

\multicolumn{1}{|l|}{\textit{\textbf{\begin{tabular}[c]{@{}l@{}} \end{tabular}}}} & \multicolumn{1}{c|}{\textbf{{$\overline{E}$}}} & \multicolumn{1}{c|}{\textbf{}} \\ \hline
\multicolumn{1}{|l|}{\textbf{\begin{tabular}[c]{@{}l@{}}HCP  pred. error\end{tabular}}} & \multicolumn{1}{c|}{0.32} & \multicolumn{1}{c|}{ 0.11    0.22    0.54} \\ \hline
\end{tabular}
\label{tab:table2}
\end{table}

The detection results from models trained on $HistD +SynthD$ and $SplitD +SynthD$ were combined and used in the post processing steps explained in Section \ref{PP} to achieve the results listed in Table \ref{tab:table2}. The proposed tracking area surround  more than half the object in close to 100 \% of the time for VB and HRS, and almost 77 \% for the SP. 

During the activities \textit{Ventilations} (BMR), \textit{Attach/remove HRS} (HRS) and  \textit{Suction} (SP) the tracking area surrounds the object during the activities in 97, 100 and  75 \% of the occurrences respectively.

Table \ref{tab:table2} also shows the results of \textit{HCP detection} and the first four results listed are estimated over all samples and episodes, and the last two results are estimated per episode.
The performance  of the detection of number of HCPs is above 90 \% when there are zero or one HCP present. However, for two and more than two HCPs the performance is 53 and 6 \% respectively. 
The mean prediction error is here 0.32, in other words, when the number of estimated HCP is incorrect, it is usually underestimated by one.  

\begin{figure}[h]
\centering
\includegraphics[width=240pt]{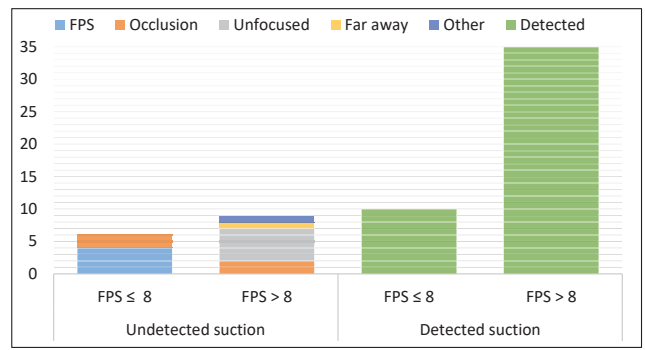}
\caption{Object detection during suction. Detected and undetected sequences with the subgroups low and medium frames per second (FPS) rate.
}  
\label{fig:SPchart}
\end{figure}

Figure \ref{fig:SPchart}  shows the distribution of the sub groups FPS $\leq 8$ and FPS $ > 8$ in the groups \textit{detected} and  \textit{undetected} SP  during suction. For the group \textit{undetected} we list the most likely reason for why the SP were undetected. The group \textit{others} represent the sequences where no large challenges was observed during the activity.

\section{Discussion}
\label{sec:discussion}

The proposed system shows promising results for object detection and tracking in noisy real-world videos  of a newborn resuscitation scene.  As proposed in Figure \ref{fig:oversikt} the areas around the objects will be used as input to sequential neural networks trained  to recognize the different activities by analyzing the areas for short time sequences. Other relevant areas like the area around the newborn, which could be found from the detected hand movements, and around the detected HCPHs can also be used as inputs to the sequential analysis. 

Due to the suction penguins transparency and small size, the system struggles with detecting it in some of the episodes. Especially in videos with low frame rate and motion blurred images it could be very difficult to detect a SP held in the hand of a health care provider. In additon, the system also has problem detecting the SP in unfocused video sequences and in activity sequences with large occlusions.
Using the sub-image approach and the \textit{SplitD} model improved the detections of the SP. This suggests  that it could be possible to further improve the results by experimenting with the size and cropping of training examples. In addition, we could experiment with the generation of the synthetic data to see if it is possible to generate more realistic examples. 

In future recordings the problem with detection of SP could be solved by using fixed camera settings, focus, frame rate and distance from resuscitation tables and by using two camera angles to avoid occlusion.    

The performance of detected number of HCPs present in the video is very good for zero and one HCP present, but the system struggles to detect the number of HCPs when there are more than one HCPs present. Instead, in cases of false detection, these are mostly being mislabeled as one HCPs less than the reference data shows. The cause for this is a mixture  of variations in the dataset and of camera angles. The system performs worse when the HCPs are not wearing rubber gloves, suggesting the need for more training examples from similar episodes. The cameras are also often placed in a side-position where the HCPs occludes other HCPs and hands.  Training the network to discriminate between left and right hands could also improve the performance of the detected number of HCPs present in the videos.

\section{Conclusion and future work}

The proposed system shows promising object detection and tracking results in noisy real-world videos. The object detection performance during activities was 97 \% on ventilation, 100  \% on attaching or removing heart rate sensor and 75 \% on suction. The system also estimate the number of health care  providers  (HCP) present with an  accuracy of 71 \%. 

In future work we will
investigate the possibility of discriminating between left and right HCP hands and implementing hand tracking to improve the performance of the estimated number of HCP.
We will also experiment with different network structures and training data to try to improve the detection of the suction device, in addition to increasing the amount of training data in general to get a better overall detection performance. 
Further, we will  continue with step two of the planned system:  inputting the proposed object areas 
to sequential neural networks to detect the resuscitation activities. 
This will produce timelines useful for quantifying the use of different resuscitation activities,  which could further provide new knowledge on the effects of activities on newborn resuscitation outcome.  
In the future, such a system could also be implemented on-site  as a   post-resuscitation debriefing tool, and/or for real-time feedback and decision support during newborn resuscitation.   The latter would require a very high-performance system. 

\label{sec:conc}

\section{Acknowledgement}

\subsection{Funding}

Our research is part of the Safer Births project which has received funding from: Laerdal Global Health, Laerdal Medical, University of Stavanger, Helse Stavanger HF, Haydom Lutheran Hospital, Laerdal Foundation for Acute Medicine, University in Oslo, University in Bergen, University of Dublin – Trinity College, Weill Cornell Medicine and Muhimbili National Hospital. The work was partly supported by the Research Council of Norway through the Global Health and Vaccination Programme (GLOBVAC) project number 228203.

For the specific study of this paper; Laerdal Medical provided the video equipment. Laerdal Global Health funded data collection in Tanzania and IT infrastructure.
The University of Stavanger funded the interpretation of the data.

\subsection{Ethical approval}

This study was approved by the National Institute of Medical Research (NIMR) in Tanzania (NIMR/HQ/R.8a/Vol. IX/1434) and the Regional Committee for Medical and Health Research Ethics (REK), Norway  (2013/110/REK vest). Parental informed verbal consent was obtained for all resuscitated newborns.

\subsection{Conflict of interests}

Myklebust is employed by Laerdal Medical. He contributed to study design and critical revision of the manuscript, but not in the analysis and interpretation of the data.

\color{black}
\bibliographystyle{ieeetran}
\bibliography{ObjDetect_refs}

\begin{thebibliography}{10}
\providecommand{\url}[1]{#1}
\csname url@samestyle\endcsname
\providecommand{\newblock}{\relax}
\providecommand{\bibinfo}[2]{#2}
\providecommand{\BIBentrySTDinterwordspacing}{\spaceskip=0pt\relax}
\providecommand{\BIBentryALTinterwordstretchfactor}{4}
\providecommand{\BIBentryALTinterwordspacing}{\spaceskip=\fontdimen2\font plus
\BIBentryALTinterwordstretchfactor\fontdimen3\font minus
  \fontdimen4\font\relax}
\providecommand{\BIBforeignlanguage}[2]{{%
\expandafter\ifx\csname l@#1\endcsname\relax
\typeout{** WARNING: IEEEtran.bst: No hyphenation pattern has been}%
\typeout{** loaded for the language `#1'. Using the pattern for}%
\typeout{** the default language instead.}%
\else
\language=\csname l@#1\endcsname
\fi
#2}}
\providecommand{\BIBdecl}{\relax}
\BIBdecl

\bibitem{birth}
S.~Wright, K.~Mathieson, L.~Brearley, S.~Jacobs, L.~Holly, R.~Wickremasinghe,
  and A.~Renton, ``Ending newborn deaths: ensuring every baby survives.'' 2014.

\bibitem{tactile}
A.~C. Lee, S.~Cousens, S.~N. Wall, S.~Niermeyer, G.~L. Darmstadt, W.~A. Carlo,
  W.~J. Keenan, Z.~A. Bhutta, C.~Gill, and J.~E. Lawn, ``Neonatal resuscitation
  and immediate newborn assessment and stimulation for the prevention of
  neonatal deaths: a systematic review, meta-analysis and delphi estimation of
  mortality effect,'' \emph{BMC public health}, vol.~11, no.~3, p. S12, 2011.

\bibitem{hege}
H.~L. Ersdal, E.~Mduma, E.~Svensen, and J.~M. Perlman, ``Early initiation of
  basic resuscitation interventions including face mask ventilation may reduce
  birth asphyxia related mortality in low-income countries: a prospective
  descriptive observational study,'' \emph{Resuscitation}, vol.~83, no.~7, pp.
  869--873, 2012.

\bibitem{huyen}
H.~Vu, K.~Engan, T.~Eftest{\o}l, A.~Katsaggelos, S.~Jatosh, S.~Kusulla,
  E.~Mduma, H.~Kidanto, and H.~Ersdal, ``Automatic classification of
  resuscitation activities on birth-asphyxiated newborns using acceleration and
  ecg signals,'' \emph{Biomedical Signal Processing and Control}, vol.~36, pp.
  20--26, 2017.

\bibitem{skaare}
C.~Sk{\aa}re, A.~M. Boldingh, J.~Kramer-Johansen, T.~E. Calisch, B.~Nakstad,
  V.~Nadkarni, T.~M. Olasveengen, and D.~E. Niles, ``Video
  performance-debriefings and ventilation-refreshers improve quality of
  neonatal resuscitation,'' \emph{Resuscitation}, vol. 132, pp. 140--146, 2018.

\bibitem{gelbart}
B.~Gelbart, R.~Hiscock, and C.~Barfield, ``Assessment of neonatal resuscitation
  performance using video recording in a perinatal centre,'' \emph{Journal of
  paediatrics and child health}, vol.~46, no. 7-8, pp. 378--383, 2010.

\bibitem{silvia}
S.~Maya-Enero, F.~Botet-Mussons, J.~Figueras-Aloy, M.~Izquierdo-Renau,
  M.~Thi{\'o}, and M.~Iriondo-Sanz, ``Adherence to the neonatal resuscitation
  algorithm for preterm infants in a tertiary hospital in spain,'' \emph{BMC
  pediatrics}, vol.~18, no.~1, p. 319, 2018.

\bibitem{nadler}
I.~Nadler, P.~M. Sanderson, C.~R. Van~Dyken, P.~G. Davis, and H.~G. Liley,
  ``Presenting video recordings of newborn resuscitations in debriefings for
  teamwork training,'' \emph{BMJ quality \& safety}, vol.~20, no.~2, pp.
  163--169, 2011.

\bibitem{kim}
K.~Schilleman, M.~L. Siew, E.~Lopriore, C.~J. Morley, F.~J. Walther, and A.~B.
  te~Pas, ``Auditing resuscitation of preterm infants at birth by recording
  video and physiological parameters,'' \emph{Resuscitation}, vol.~83, no.~9,
  pp. 1135--1139, 2012.

\bibitem{conventional}
J.~D. Courtney, ``Automatic video indexing via object motion analysis,''
  \emph{Pattern Recognition}, vol.~30, no.~4, pp. 607--625, 1997.

\bibitem{activityDetection1}
G.~Gkioxari and J.~Malik, ``Finding action tubes,'' in \emph{Proceedings of the
  IEEE conference on computer vision and pattern recognition}, 2015, pp.
  759--768.

\bibitem{activityDetection2}
S.~Ma, L.~Sigal, and S.~Sclaroff, ``Learning activity progression in lstms for
  activity detection and early detection,'' in \emph{Proceedings of the IEEE
  Conference on Computer Vision and Pattern Recognition}, 2016, pp. 1942--1950.

\bibitem{activityDetection3}
A.~Montes, A.~Salvador, S.~Pascual, and X.~Giro-i Nieto, ``Temporal activity
  detection in untrimmed videos with recurrent neural networks,'' \emph{arXiv
  preprint arXiv:1608.08128}, 2016.

\bibitem{activityDetection4}
B.~Singh, T.~K. Marks, M.~Jones, O.~Tuzel, and M.~Shao, ``A multi-stream
  bi-directional recurrent neural network for fine-grained action detection,''
  in \emph{Proceedings of the IEEE Conference on Computer Vision and Pattern
  Recognition}, 2016, pp. 1961--1970.

\bibitem{RFID1}
S.~Parlak and I.~Marsic, ``Detecting object motion using passive rfid: A trauma
  resuscitation case study,'' \emph{IEEE Transactions on Instrumentation and
  Measurement}, vol.~62, no.~9, pp. 2430--2437, 2013.

\bibitem{RFID2}
S.~Parlak, A.~Sarcevic, I.~Marsic, and R.~S. Burd, ``Introducing rfid
  technology in dynamic and time-critical medical settings: Requirements and
  challenges,'' \emph{Journal of biomedical informatics}, vol.~45, no.~5, pp.
  958--974, 2012.

\bibitem{RFID3}
S.~Parlak, I.~Marsic, A.~Sarcevic, W.~U. Bajwa, L.~J. Waterhouse, and R.~S.
  Burd, ``Passive rfid for object and use detection during trauma
  resuscitation,'' \emph{IEEE Transactions on Mobile Computing}, vol.~15,
  no.~4, pp. 924--937, 2016.

\bibitem{ActivityRecognitionResuscitation}
I.~Chakraborty, A.~Elgammal, and R.~S. Burd, ``Video based activity recognition
  in trauma resuscitation,'' in \emph{2013 10th IEEE International Conference
  and Workshops on Automatic Face and Gesture Recognition (FG)}.\hskip 1em plus
  0.5em minus 0.4em\relax IEEE, 2013, pp. 1--8.

\bibitem{activityDetectionNewbornVideos}
Y.~Guo, J.~Wrammert, K.~Singh, K.~Ashish, K.~Bradford, and A.~Krishnamurthy,
  ``Automatic analysis of neonatal video data to evaluate resuscitation
  performance,'' in \emph{Computational Advances in Bio and Medical Sciences
  (ICCABS), 2016 IEEE 6th International Conference on}.\hskip 1em plus 0.5em
  minus 0.4em\relax IEEE, 2016, pp. 1--6.

\bibitem{hist}
R.~C. Gonzalez and R.~E. Woods, \emph{Digital Image Processing}, third
  edition~ed.\hskip 1em plus 0.5em minus 0.4em\relax Pearson, 2008.

\bibitem{YOLOv3}
J.~Redmon and A.~Farhadi, ``Yolov3: An incremental improvement,'' \emph{arXiv},
  2018.

\bibitem{yolo1}
W.~Xu and S.~Matzner, ``Underwater fish detection using deep learning for water
  power applications,'' \emph{arXiv preprint arXiv:1811.01494}, 2018.

\bibitem{yolo2}
L.~Heng, B.~Choi, Z.~Cui, M.~Geppert, S.~Hu, B.~Kuan, P.~Liu, R.~Nguyen, Y.~C.
  Yeo, A.~Geiger \emph{et~al.}, ``Project autovision: Localization and 3d scene
  perception for an autonomous vehicle with a multi-camera system,''
  \emph{arXiv preprint arXiv:1809.05477}, 2018.

\bibitem{yolo3}
I.~S{\'a}r{\'a}ndi, T.~Linder, K.~O. Arras, and B.~Leibe, ``Synthetic occlusion
  augmentation with volumetric heatmaps for the 2018 eccv posetrack challenge
  on 3d human pose estimation,'' \emph{arXiv preprint arXiv:1809.04987}, 2018.

\bibitem{huyenLNRM}
H.~Vu, T.~Eftest{\o}l, K.~Engan, J.~Eilevstj{\o}nn, L.~B. Yarrot, J.~E. Linde,
  and H.~L. Ersdal, ``Automatic detection and parameterization of manual
  bag-mask ventilation on newborns,'' \emph{IEEE journal of biomedical and
  health informatics}, vol.~21, no.~2, pp. 527--538, 2017.

\bibitem{label}
MATLAB and I.~L. Computer Vision System~Toolbox, ``The mathworks, inc., natick,
  massachusetts, united states.''

\bibitem{residual}
K.~He, X.~Zhang, S.~Ren, and J.~Sun, ``Deep residual learning for image
  recognition,'' in \emph{Proceedings of the IEEE conference on computer vision
  and pattern recognition}, 2016, pp. 770--778.

\bibitem{pyramid}
T.-Y. Lin, P.~Doll{\'a}r, R.~B. Girshick, K.~He, B.~Hariharan, and S.~J.
  Belongie, ``Feature pyramid networks for object detection.'' in \emph{CVPR},
  vol.~1, no.~2, 2017, p.~4.

\bibitem{MOTAsimple}
L.~{\v{C}}ehovin, A.~Leonardis, and M.~Kristan, ``Visual object tracking
  performance measures revisited,'' \emph{IEEE Transactions on Image
  Processing}, vol.~25, no.~3, pp. 1261--1274, 2016.

\end{thebibliography}
\end{document}